\documentclass[letterpaper, 10 pt, conference]{class/ieeeconf}  % Comment this line out
\IEEEoverridecommandlockouts                              % This command is only
\usepackage{xcolor}
\usepackage{siunitx}  

\usepackage[algo2e]{algorithm2e}
\usepackage{graphicx}
 \graphicspath{./figures/robot_demonstration}
\usepackage{amsmath}
\usepackage{amssymb}
\usepackage{latexsym}
\usepackage{url}
\usepackage{cite}
\usepackage{relsize}
\usepackage{multirow}
\usepackage{afterpage}
\usepackage{ifthen}
\usepackage{graphicx}
\usepackage{algpseudocode,algorithm,algorithmicx}
\usepackage{subfigure}
\usepackage{flushend}
\usepackage{epstopdf}
\usepackage{stackengine}
\usepackage{textcomp} % new

\usepackage{gensymb}
\usepackage{booktabs}
\usepackage{makecell}
\usepackage{hhline}
\usepackage{courier}
\usepackage{lipsum}

\usepackage{xcolor} % new
            
\makeatletter
\let\NAT@parse\undefined
\makeatother
\usepackage[bookmarks=false, linkcolor=blue, urlcolor=blue, citecolor=blue]{hyperref} 

\hypersetup{
    colorlinks=true,
    linkcolor=red,
    filecolor=magenta,      
    urlcolor=blue,
    pdfstartview={FitH},
    citecolor =blue
    }

\usepackage{xspace}

\title{\LARGE \bf Lightweight Language-driven Grasp Detection using \\ Conditional Consistency Model}

\author{Nghia Nguyen$^{1}$, Minh Nhat Vu$^{2,3,*}$, Baoru Huang$^4$, An Vuong$^{1}$, Ngan Le$^5$, Thieu Vo$^6$, Anh Nguyen$^7$
\thanks{$^1$ FPT Software AI Center, Vietnam {\tt nghiant100@fpt.com}}
\thanks{$^2$ Automation \& Control Institute, TU Wien, Vienna, Austria %{\tt vu@acin.tuwien.ac.at}
}
\thanks{$^3$ Austrian Institute of Technology (AIT) GmbH, Austria %{\tt vu@acin.tuwien.ac.at}
}
\thanks{$^4$ Imperial College London, UK
}
\thanks{$^5$ University of Arkansas, USA 
}
\thanks{$^6$ National University of Singapore, Singapore} %{\tt vongocthieu@tdtu.edu.vn}}
\thanks{$^7$ Department of Computer Science, University of Liverpool, UK %{\tt anh.nguyen@liverpool.ac.uk}
}
\thanks{$^*$ Corresponding author {\tt minh.vu@ait.ac.at}
}
}

\begin{document}
% Macros

\newtheorem{problem}{Problem}
\newtheorem{lemma}{Lemma}
\newtheorem{theorem}[lemma]{Theorem}
\newtheorem{claim}{Claim}
\newtheorem{corollary}[lemma]{Corollary}
\newtheorem{definition}[lemma]{Definition}
\newtheorem{proposition}[lemma]{Proposition}
\newtheorem{remark}[lemma]{Remark}
\newenvironment{LabeledProof}[1]{\noindent{\it Proof of #1: }}{\qed}

\def\beq#1\eeq{\begin{equation}#1\end{equation}}
\def\bea#1\eea{\begin{align}#1\end{align}}
\def\beg#1\eeg{\begin{gather}#1\end{gather}}
\def\beqs#1\eeqs{\begin{equation*}#1\end{equation*}}
\def\beas#1\eeas{\begin{align*}#1\end{align*}}
\def\begs#1\eegs{\begin{gather*}#1\end{gather*}}

\newcommand{\poly}{\mathrm{poly}}
\newcommand{\eps}{\epsilon}
\newcommand{\e}{\epsilon}
\newcommand{\polylog}{\mathrm{polylog}}
\newcommand{\rob}[1]{\left( #1 \right)} %Round Brackets
\newcommand{\sqb}[1]{\left[ #1 \right]} %square Brackets
\newcommand{\cub}[1]{\left\{ #1 \right\} } %curly brackets
\newcommand{\rb}[1]{\left( #1 \right)} %Round
\newcommand{\abs}[1]{\left| #1 \right|} %| |
\newcommand{\zo}{\{0, 1\}}
\newcommand{\zonzo}{\zo^n \to \zo}
\newcommand{\zokzo}{\zo^k \to \zo}
\newcommand{\zot}{\{0,1,2\}}
\newcommand{\en}[1]{\marginpar{\textbf{#1}}}
\newcommand{\efn}[1]{\footnote{\textbf{#1}}}
\newcommand{\vecbm}[1]{\boldmath{#1}} %more general (handles greek letters)
\newcommand{\uvec}[1]{\hat{\vec{#1}}}
\newcommand{\thv}{\vecbm{\theta}}
\newcommand{\junk}[1]{}
\newcommand{\var}{\mathop{\mathrm{var}}}
\newcommand{\rank}{\mathop{\mathrm{rank}}}
\newcommand{\diag}{\mathop{\mathrm{diag}}}
\newcommand{\tr}{\mathop{\mathrm{tr}}}
\newcommand{\acos}{\mathop{\mathrm{acos}}}
\newcommand{\atantwo}{\mathop{\mathrm{atan2}}}
\newcommand{\SVD}{\mathop{\mathrm{SVD}}}
\newcommand{\quadf}{\mathop{\mathrm{q}}}
\newcommand{\linterp}{\mathop{\mathrm{l}}}
\newcommand{\sgn}{\mathop{\mathrm{sign}}}
\newcommand{\sym}{\mathop{\mathrm{sym}}}
\newcommand{\avg}{\mathop{\mathrm{avg}}}
\newcommand{\mean}{\mathop{\mathrm{mean}}}
\newcommand{\erf}{\mathop{\mathrm{erf}}}
\newcommand{\grad}{\nabla}
\newcommand{\R}{\mathbb{R}}
\newcommand{\defeq}{\triangleq}
\newcommand{\dims}[2]{[#1\!\times\!#2]}
\newcommand{\sdims}[2]{\mathsmaller{#1\!\times\!#2}}
\newcommand{\udims}[3]{#1}
\newcommand{\udimst}[4]{#1}
\newcommand{\com}[1]{\rhd\text{\emph{#1}}}
\newcommand{\ind}{\hspace{1em}}
\newcommand{\argmin}[1]{\underset{#1}{\operatorname{argmin}}}
\newcommand{\floor}[1]{\left\lfloor{#1}\right\rfloor}
\newcommand{\step}[1]{\vspace{0.5em}\noindent{#1}}
\newcommand{\quat}[1]{\ensuremath{\mathring{\mathbf{#1}}}}
\newcommand{\norm}[1]{\left\lVert#1\right\rVert}
\newcommand{\ignore}[1]{}
\newcommand{\specialcell}[2][c]{\begin{tabular}[#1]{@{}c@{}}#2\end{tabular}}
\newcommand*\Let[2]{\State #1 $\gets$ #2}
\newcommand{\algorithmicbreak}{\textbf{break}}
\newcommand{\Break}{\State \algorithmicbreak}
\newcommand{\ra}[1]{\renewcommand{\arraystretch}{#1}}

\renewcommand{\vec}[1]{\mathbf{#1}} %looks better

\algdef{S}[FOR]{ForEach}[1]{\algorithmicforeach\ #1\ \algorithmicdo}
\algnewcommand\algorithmicforeach{\textbf{for each}}
\algrenewcommand\algorithmicrequire{\textbf{Require:}}
\algrenewcommand\algorithmicensure{\textbf{Ensure:}}
\algnewcommand\algorithmicinput{\textbf{Input:}}
\algnewcommand\INPUT{\item[\algorithmicinput]}
\algnewcommand\algorithmicoutput{\textbf{Output:}}
\algnewcommand\OUTPUT{\item[\algorithmicoutput]}

\maketitle
\thispagestyle{empty}
\pagestyle{empty}

%%%%%%%%%%%%%%%%%%%%%%%%%%%%%%%%%%%%%%%%%%%%%%%%%%%%%%%%%%%%%%%%%%%%%%%%%%%%%%%%
\begin{abstract}
Language-driven grasp detection is a fundamental yet challenging task in robotics with various industrial applications. In this work, we present a new approach for language-driven grasp detection that leverages the concept of lightweight diffusion models to achieve fast inference time. By integrating diffusion processes with grasping prompts in natural language, our method can effectively encode visual and textual information, enabling more accurate and versatile grasp positioning that aligns well with the text query. To overcome the long inference time problem in diffusion models, we leverage the image and text features as the condition in the consistency model to reduce the number of denoising timesteps during inference. The intensive experimental results show that our method outperforms other recent grasp detection methods and lightweight diffusion models by a clear margin. We further validate our method in real-world robotic experiments to demonstrate its fast inference time capability.

%demonstrate the efficacy of our method in various scenarios, outperforming existing methods in terms of accuracy and resource usage. 

%The proposed approach paves the way for more intuitive and accessible robotic grasping technologies, holding promise for vast improvements in automated systems and human-robot interaction.
\end{abstract}

%%%%%%%%%%%%%%%%%%%%%%%%%%%%%%%%%%%%%%%%%%%%%%%%%%%%%%%%%%%%%%%%%%%%%%%%%%%%%%%%

\section{INTRODUCTION} \label{Sec:Intro}
%DO NOT DELETE \note{figure link: shorturl.at/glALZ} 

% Create an Introduction Figure !!!

Grasping is one of the fundamental tasks in robotics, enabling robots to interact with the physical world through a broad spectrum of applications, from industrial automation and human-robot interaction to service robotics~\cite{wang2021double}. Recent advancements in machine vision have significantly improved the capabilities of grasp detection for the robot~\cite{sundermeyer2021contacT,wen2022catgrasp, ainetter2021end,depierre2018jacquard,shridhar2022cliport}. Prior research has demonstrated encouraging grasp detection results in both 2D images~\cite{ainetter2021end,chu2018real} and 3D point clouds~\cite{nguyen2023open, alliegro2022end}. However, most existing works define grasp detection as a region localization problem while ignoring the use of natural language to localize possible grasps on the object based on linguistic input~\cite{chen2021joint}. 

%In practice, using language in manipulation tasks facilitates efficient communication and bridging the gap between human intent and robot execution~\cite{yang2023pave}.

%ignore the    these results are only applicable to small datasets, limited to serving daily routine tasks. Additionally, these methods have yet to address the issue of out of vocabulary problems.

With the recent advances in Large Language Models (LLM), integrating language to robotic systems has become more popular~\cite{vuong2023open}. Pretrained models such as ChatGPT~\cite{radford2018improving} and CLIP~\cite{radford2021learning} have revolutionized various applications and their adaptability to the robotic domain has shown encouraging results~\cite{tziafas2023language,xu2023joint, song2023learning,yang2023pave}. Although there are several language-driven robotic manipulations works, most focus on understanding high-level actions and overlook the fundamental grasping task~\cite{brohan2023can}. In this paper, we tackle the \textit{language-driven grasp detection} task that allows the robot to grasp specific objects based on the language command. With the language-driven grasping ability, the robot would be able to interact more effectively with the surrounding environment and humans.

% \begin{figure}[t]
% \centering
% \includegraphics[height=50mm]{figures/intro/introduce.pdf}
% \vspace{2ex}
% \caption{The comparison between  (a) traditional grasp detection methods without language, and (b) our Lightweight Language-driven Grasp Detection (LLGD) that can determine the grasp poses based on language input.}
% \label{fig:single_intro}
% \end{figure}

% \begin{figure}[htbp]
%   \centering
  
%   \begin{subfigure}[b]{0.45\textwidth}    % [b] là căn chỉnh dưới cùng, 0.45\textwidth là chiều rộng của subfigure
%     \centering
%     \includegraphics[width=\textwidth]{figures/intro/introduce_1.pdf}    % Thay 'image1.jpg' bằng tên file ảnh thực tế của bạn
%     \caption{(a)}
%     \label{fig:sub1}
%   \end{subfigure}
%   \hfill  % Điều chỉnh khoảng cách giữa các subfigure
%   \begin{subfigure}[b]{0.45\textwidth}    % Lặp lại cho subfigure thứ hai
%     \centering
%     \includegraphics[width=\textwidth]{figures/intro/introduce_2.pdf}    % Thay 'image2.jpg' bằng tên file ảnh thực tế của bạn
%     \caption{(b)}
%     \label{fig:sub2}
%   \end{subfigure}
  
%   \caption{Tiêu đề chung cho cả figure}
%   \label{fig:combined}
% \end{figure}

Compared to the traditional grasp detection task without text, language-driven grasping offers several advantages. Firstly, we communicate with robots by providing language prompts that direct them to execute precise tasks~\cite{chen2021joint,tziafas2023language,xu2023joint, song2023learning,yang2023pave,shridhar2022cliport}; therefore, the incorporation of natural language instructions augments robotic systems with the ability to interactively respond to dynamic, real-time tasks~\cite{lynch2023interactive}. Secondly, the utilization of natural language addresses the challenge of ambiguity in identifying target objects within cluttered environments~\cite{lu2023vl} or distinguishing among objects with similar shapes~\cite{ma2023towards}. Lastly, linguistic guidance enriches robotic systems with semantic information~\cite{shah2023lm}, enhancing their learning capabilities without necessitating expert demonstrations or specific engineering~\cite{ha2023scaling}.

% \textbf{write at here} \lipsum[2].

Recently, several works on grasp detection have utilized diffusion models as the key technique and shown encouraging results~\cite{chen2023diffusiondet, xu2023joint, urain2023se}. This is motivated by the proven efficacy of diffusion models in conditional generation tasks~\cite{ho2020denoising} such as image synthesis, image segmentation, and visual grounding~\cite{chen2023diffusiondet}. The effectiveness of diffusion models comes from their iterative approach to gradually refine data from an initial state of pure noise toward a meaningful output. Nonetheless, applying diffusion models to language-driven tasks in robotics faces a key challenge, \textit{i.e.}, the inference time of diffusion models is usually not fast enough for real-time robotic applications. Consequently, recent studies have introduced techniques to tackle the inference speed problem of diffusion models using approaches such as rapid sampling~\cite{li2023snapfusion, salimans2022progressive,song2023consistency}, knowledge distillation~\cite{habibian2023clockwork, song2023consistency}, or model optimization~\cite{chen2023lightgrad,liu2023rapid}. However, these models are still unable to perform fast sampling with language conditions during inference to meet the real-time requirement in robotic grasping.

%that retain the efficacy of their larger counterparts. The development of these streamlined models has been instrumental in expanding the applicability of Diffusion Model technology to resource-constrained platforms. 

In this paper, we propose a new lightweight diffusion model to tackle the inference speed problem in utilizing the diffusion model for the language-driven grasp detection task. To this end, we exploit the capabilities of flow-based generative models to improve the precision of robots in identifying grasp poses from textual inputs. In particular, we develop a conditional consistency model for fast inference speed for real-time robotic applications. We verify our proposed method on a recent large-scale language-driven grasping dataset and achieve superior results in both accuracy and inference speed compared with recent approaches. Furthermore, our method enables zero-shot learning and generalize to real-world robotic grasping applications.

 % Our model is developed based on the recent consistency model~\cite{song2023consistency} with two distinct new features: \textit{i)} the integration of \textit{condition} to consistent model for the lang output for stable results \textit{i)}

Our contributions are summarized as follows:
\begin{itemize}
    \item 
    We present Lightweight Language-driven Grasp Detection (LLGD), a fast diffusion model for language-driven grasp detection.
    %\item We transform Diffusion Models into an equivalent variant - \textbf{Consistency Models}~\cite{song2023consistency}, while also proposing a training objective for the problem.
    \item 
    We conduct intensive analysis to validate our method and demonstrate that it outperforms other approaches in terms of both accuracy and execution speed.
    
\end{itemize}

\section{Related Work} \label{Sec:rw}

%Affordance detection is one of the most important task among several tasks involved in the large field of affordance understanding since it concerns not only the available affordances of objects but also the potential regions where these affordances are provided. With the ever-increasing improvement in the performance of convolutional neural networks, several CNN-based approaches have been proposed to tackle affordance detection on images~\cite{hassanin2021visual}.

\textbf{Grasp Detection.} Grasp detection has been a central topic in robotics, aiming to equip robots with the ability to identify and execute object grasping in complex environments~\cite{ainetter2021end, depierre2018jacquard, shridhar2022cliport, vuong2023grasp, dai2023graspnerf, chen2023keypoint}.  Several works such as that by Redmon \textit{et al.}~\cite{redmon2015real} have set the foundation for robot grasping by using convolutional neural networks (CNNs). Most previous grasp detection methods are often limited to simple tasks with a fixed number of classes and rely solely on raw image data~\cite{redmon2015real,depierre2018jacquard,chu2018real}. Several works~\cite{kumra2017robotic,ni2020pointnet++,alliegro2022end} have extended the problem by using RGB-D images or 3D point clouds to output the results in 3D space. However, they still have not focused on integrating language as the input instruction in the grasp detection problem. % able to address the out-of-vocabulary (OOV) issue. As a result, recent studies have developed robotic systems with the ability to follow natural commands.

\textbf{Language-driven Grasping.} Language-driven grasp detection introduces the use of natural language to inform grasp detection tasks~\cite{shridhar2022cliport,nguyen2024LGrasp6D,xu2023joint,vuong2024language,tuan2024seggrasp}. The common approach to tackling the task of language-driven grasp detection is to divide it into a two-step process. One stage is dedicated to identifying the target object, and the second stage focuses on generating grasp poses based on the established visual-text correlations~\cite{vuong2023grasp}. Foundation models such as GroundDINO~\cite{liu2023grounding}, CLIP~\cite{radford2021learning} have emerged, enabling zero-shot detection and zero-shot segmentation. These models allow for the localization of the target object without training~\cite{yang2023pave}. However, due to their large size, they result in longer inference times. Accessing such commercial foundation models is not always possible, especially since LLM models often require the use of APIs which come at a high cost.

\textbf{Lightweight Diffusion Model.} Lightweight diffusion models that maintain performance while reducing computational overhead have become crucial in machine learning. %Li \textit{et al.}~\cite{li2023snapfusion} demonstrates how pruning U-Net and using step distillation~\cite{salimans2022progressive} can effectively reduce model size and complexity in diffusion models. %Chen \textit{et al.}~\cite{chen2023lightgrad} designed a lightweight U-net model and employed a fast sampling technique in the continuous domain.
Habibian \textit{et al.}~\cite{habibian2023clockwork} utilized knowledge distillation for low-resolution features to reduce the number of parameters in U-Net. Song \textit{et al.}~\cite{song2020score} introduced the concept of score-based generative models. Recently, consistency models have surfaced as a strong approach of generative models capable of producing high-quality images within a single or a limited number of steps~\cite{song2023consistency}. Although there are significant applications in generative tasks, these models are mostly \textit{unconditional}~\cite{song2023consistency,salimans2022progressive,chen2023lightgrad}. %Objective functions are essential in these frameworks to assess the correlation between input and generated outputs. For example, CLIP score~\cite{radford2021learning} for text-to-image, MCD score~\cite{chen2023lightgrad} for text-to-speech. Unfortunately, there is still no equivalent function for grasp detection problem. 
On the other hand, robotic applications remain discriminative, making the use of unconditional diffusion models not entirely suitable. In this study, we address this issue by building a lightweight diffusion with \textit{language conditions}. We aim to enhance the consistency model work~\cite{song2023consistency} to inherit its fast inference time while adding the language conditions to make it more suitable for the language-driven grasping task.

%among lightweight models, by adapting it into a corresponding conditional form.

%From there, he identified the Probability Flow ODE (PF ODE) trajectory of a complex diffusion process. Instead of performing the denoising process step by step like Diffusion Models, Consistency Models learn a mapping from arbitrary noise and time-step to obtain the output directly. Therefore, Consistency Models are capable of generating samples in a few steps while still yielding results comparable to Diffusion Models. This is key to reduce the inference time of our lightweight grasp detection model.

%Language-driven segmentation has received research interest from computer vision and machine learning recently. 
% Language-driven segmentation has recently attracted research interest in computer vision and machine learning.
%Most recent works are inspired by the desirable result of large-scale language models such as CLIP~\cite{radford2021learning} or BERT~\cite{devlin2018bert}. 

%impressive results achieved in semantic segmentation, the potential exploitation of the language guidance for the task of point-wise affordance detection has yet to be thoroughly investigated. This work, to the best of our knowledge, is the first attempt to apply the concept of language-driven for the open-vocabulary affordance detection task in 3D point clouds.}

% \textbf{Consistency Model. }  
\section{Lightweight Language-driven Grasp Detection} \label{Sec:method}

% \begin{figure*}[t]
% 	\centering
% 	\includegraphics[width=0.85\linewidth]{figures/method/MPC_Grasp.png}
%  \vspace{1pt}
% 	\caption{The overview of the proposed Open-Vocabulary Affordance Detection (OpenAD) network. 
%                  First, the input point cloud is fed into a point cloud network to extract per-point embeddings. 
%                  Second, the affordance labels are passed into a text encoder to extract the text embeddings. 
%                  %Next, the point-wise feature and the text embedding are correlated using cosine similarity. 
%                  Subsequently, the correlation between the point-wise features and the corresponding text embeddings is computed using the cosine similarity function. 
%                  Finally, a softmax layer is employed to predict language-driven affordances.}
% 	\label{fig:architecture}
% \end{figure*}
\subsection{Overview}
Given an input RGB image and a text prompt describing the object of interest, we aim to detect the grasping pose on the image that best matches the text prompt input. We follow the popular \textit{rectangle grasp} convention widely used in previous work to define the grasp~\cite{depierre2018jacquard,kumra2020antipodal,ainetter2021end}.
We represent the target grasp pose as $\mathbf{x}_0$ in the diffusion model. The objective of our diffusion process of language-driven grasp detection involves denoising from a noisy state $\mathbf{x}_T$ to the original grasp pose $\mathbf{x}_0$, conditioned on the input image and grasp instruction represented by $y$. The forward process in traditional conditional diffusion model~\cite{ho2020denoising} is defined as:
\begin{equation}
    q(\mathbf{x}_t|\mathbf{x}_{t-1}) = \mathcal{N}(\sqrt{1-\beta_t}\mathbf{x}_{t-1},\beta_t\mathbf{I})~,
\end{equation}
where the hyperparameter $\beta_t$ is the amount of noise added at diffusion step $t \in [0,T] \subset \mathbb{R}$. %Each step depends only on the previous one, allowing us to derive a closed-form formula:
% \begin{equation}
%     \label{eq:forward_discrete_process}
%     p_t(\mathbf{x}_t|\mathbf{x}_0) = \mathcal{N}(\sqrt{\bar{\alpha}_t}\mathbf{x}_0, (1 - \bar{\alpha}_t)\mathbf{I})
% \end{equation}

To train a diffusion model with condition $y$, we use a neural network to learn the reverse process:
\begin{equation}
    \label{eq:discrete_diffusion}
    p_\phi(\mathbf{x}_{t-1}|\mathbf{x}_t,y) = \mathcal{N}(\mu_\phi(\mathbf{x}_t,t,y),\Sigma_\phi(\mathbf{x}_t,t,y))~.
\end{equation}

In our approach, we utilize the diffusion process in the continuous domain, where $\mathbf{x}_t$ is the grasp pose state at arbitrary time index $t$. %We recall that we represent the predicted grasp pose as $\mathbf{x}_0$, while $\mathbf{x}_T$ represents a noisy state. 
Unlike popular discrete diffusion models as in previous studies~\cite{chen2023diffusiondet,le2023controllable,zhang2022time,ho2020denoising,vuong2024nips}, by using a continuous space, we can improve sample quality and reduce inference times due to the ability to traverse the diffusion process at arbitrary timesteps, allowing for more fine-grained control over the denoising process~\cite{chen2023lightgrad}. 

%Chen \textit{et al.}~\cite{chen2023lightgrad} introduced a fast sampling technique using Continuous Diffusion Models (CDMs) that effectively reduces the number of iterations to an integer factor. CDMs can leverage advanced numerical methods for solving stochastic differential equations (SDEs)~\cite{van1976stochastic}.

\begin{figure*}[t] 
   \centering
   \includegraphics[width=0.95\linewidth]{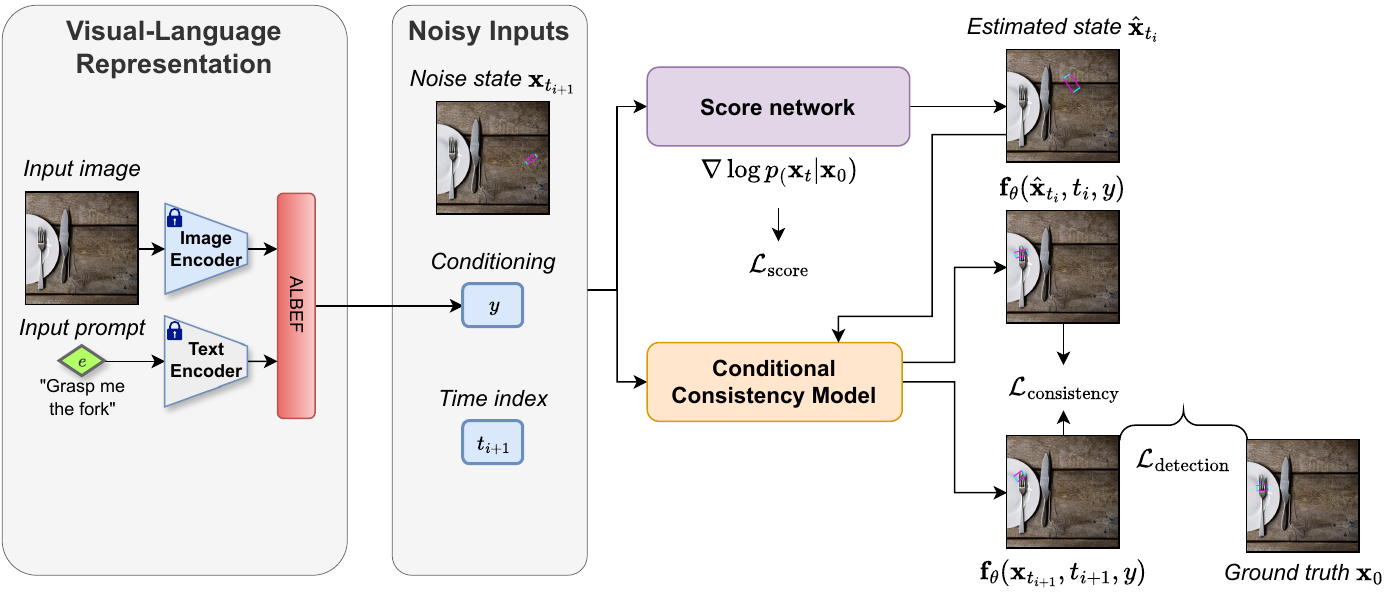} 
   \caption{The overview of our method. First, the input RGB image and text prompt are fed into the feature encoder and ALBEF fusion~\cite{li2021align}. Subsequently, we concurrently train two models with the same architectures: A score network to estimate the probability flow Ordinary Differential Equation (ODE) trajectory~\cite{song2020score} for the diffusion process and a conditional consistency model to determine the grasp pose with a few denoising steps.}
   \label{fig:lightgrasping}
\end{figure*}

\subsection{Conditional Consistency Model for LLGD}

%From Equation~\ref{eq:discrete_diffusion}, we need to implement diffusion process repetitively. This causes the diffusion process to proceed slowly, especially with large values of $T$. Similar to PFGM~\cite{xu2022poisson},

To reduce the inference time during the denoising step of the diffusion model, we aim to estimate the original grasp pose with just a few denoising steps. Since our language-driven grasp detection task has the condition $y$, we introduce a \textit{conditional} consistency model based on the consistency concept in~\cite{song2023consistency} to infer the original grasp pose during the inference process directly:

% a single step to reduce the latency of the inference process. We introduce a novel neural network that employs the Consistency Model concept

% \begin{equation}
%     p_\theta(\mathbf{x}_0|\mathbf{x_t},\mathbf{r}) 
% \end{equation}

\begin{equation}
    \mathbf{f}_\theta(\mathbf{x}_t,t,y) = 
    \begin{cases}
    \mathbf{x}_t & t \in [0,\epsilon] \\
    \mathbf{F}_\theta(\mathbf{x}_t,t,y) & t \in (\epsilon,T]
    \end{cases}~,
\end{equation}
where $\mathbf{f}_\theta(\mathbf{x}_\epsilon, t, y) = \mathbf{x}_\epsilon$ is the boundary condition, $\mathbf{F}_\theta(\mathbf{x}_t,t,y)$ is a free-form deep neural network whose output has same dimentionality as $\mathbf{x}_t$. 

% From~\cite{chen2023lightgrad}, we can reduce the latency for diffusion model using fast sampling technique. 
% In the core section of the methodology, we utilize a Flow-based Generative Model for the grasp detection problem. Before proceeding to generate the grasp pose, we declare the Vision-Language Representation $\mathbf{r}$. To clearly analyze the ways to reduce the latency for the model, we define the forward
% process and backward process of LLGD in terms of SDE.
To train our conditional consistency model, we employ knowledge distillation from a continuous diffusion process: % with a stochastic differential equation (SDE):

\begin{equation}
    \label{eq:sde_equation}
    d\mathbf{x}_{t} = -\frac{1}{2}\gamma_t\mathbf{x}_tdt + \sqrt{\gamma_t}d\mathbf{w}_t~,
\end{equation}
where $\gamma_t$ is non-negative function referred as noise schedule, $\mathbf{w}_t$ is the standard Brownian motion~\cite{chen2023lightgrad}. This forward process creates a trajectory of grasp pose $\{\mathbf{x}_t\}_{t=0}^T$. The grasp pose state $\mathbf{x}_t$ is not only dependent on the time index $t$ but also on the input image and text prompt. The grasp distribution $p(\mathbf{x}_0|y)$ from dataset is transformed into $p(\mathbf{x}_T|y)\sim\mathcal{N}(0,\mathbf{I})$. Given ground truth grasp pose $\mathbf{x}_0$, we can sample $\mathbf{x}_t$ at arbitrary $t$:
\begin{equation}
    \label{eq:distribution_gaussian}
    p(\mathbf{x}_t|\mathbf{x}_0) = \mathcal{N}(\mu_t,\Sigma_t)~,
\end{equation}
where 
$$
    \label{eq:parameter_add_noise}
    \mu_t = e^{\frac{1}{2}\rho_t}\mathbf{x}_0,
    \Sigma_t = (1 - e^{\rho_t})\mathbf{I}, \rho_t = -\int_{0}^{t}{\gamma_sds}~.
$$

Equation~\ref{eq:sde_equation} is a probability flow ODE~\cite{song2020score}. With the conditional variable $y$, it can be redefined as:
\begin{equation}
    \label{eq:PF_ODE}
    \frac{d\mathbf{x}_t}{dt} = -\frac{1}{2}\gamma_t\left[\mathbf{x}_t + \nabla\log p(\mathbf{x}_t|y)\right]~,
\end{equation}
where $\nabla\log p(\mathbf{x}_t|y)$ is score function of conditional diffusion model.

% Using Bayes's Rule for $p_t(\mathbf{x}_t|\mathbf{r})$, we have:
% \begin{equation}
%     p_t(\mathbf{x}_t|\mathbf{r}) = \frac{p_t(\mathbf{r}|\mathbf{x}_t)p_t(\mathbf{x}_t)}{p_{data}(\mathbf{r})}
% \end{equation}
% where $p_{data}(\mathbf{r})$ is image and text prompt data distribution, independent with $\mathbf{x}_t$. It is not affect to score function:
% \begin{equation}
%     \nabla \log p_t(\mathbf{x}_t|\mathbf{r}) = \nabla \log p_t(\mathbf{r}|\mathbf{x}_t) + \nabla \log p_t(\mathbf{x}_t)
% \end{equation}

% \textit{Remark 1:} Gaussian distribution in continuous domain in Eq~\ref{eq:distribution_gaussian} have properties similar with discrete diffusion models in Eq~\ref{eq:forward_discrete_process}. In this case, if we choose $\bar{\alpha}_t = e^{\rho_t}$, we can convert arbitrary Diffusion Models in the discrete domain to a continuous form and vice versa. However, the inference process for different models always requires setting the step size to 1, while in the continuous form, we can choose an appropriate sampling time to speed up inference.

Suppose that we have a neural network $\mathbf{s}_\phi(\mathbf{x}_t,t,y)$ that can approximate the score function $\nabla\log p(\mathbf{x}_t|y)$, \textit{i.e.}, $\mathbf{s}_\phi(\mathbf{x}_t,t,y) \approx \nabla\log p(\mathbf{x}_t|y)$, after training the score network, we can replace the $\nabla\log p(\mathbf{x}_t|y)$ term in Equation~\ref{eq:PF_ODE} with a neural network:
\begin{equation}
    \label{eq:approximated_PFODE}
     \frac{d\mathbf{x}_t}{dt} = -\frac{1}{2}\gamma_t\left[\mathbf{x}_t + \mathbf{s}_\phi(\mathbf{x}_t,t,y)\right]~.
\end{equation}

\textbf{Score Function Loss. }In order to approximate score function $\nabla\log p(\mathbf{x}_t|y)$, the conditional denoising estimator minimizes following objective:

\begin{equation}
    \begin{aligned}
    \mathcal{L}_{\rm score}=\mathbb{E}_{
        \begin{subarray}{l}
        t \sim \mathcal{U}[0, T] \\
        \mathbf{x}_0,y \sim p(\mathbf{x}_0,y) \\
        \mathbf{x}_t \sim p(\mathbf{x}_t|\mathbf{x}_0)
        \end{subarray}
        }\left[\lambda(t) \|\nabla\log p(\mathbf{x}_t|\mathbf{x}_0) - \mathbf{s}_\phi(\mathbf{x}_t,t,y)\|^2 \right]~,
    \end{aligned}
\end{equation}
where $\lambda(t) \in \mathbb{R}^+$ is a positive weighting function.

\textbf{Proposition 1. } \textit{Suppose that $\mathbf{x}_t$ is conditionally independent of $y$ given $\mathbf{x}_0$, then minimizing of $\mathcal{L}_{\rm score}$ is the same as minimizing:}
$$
    \mathbb{E}_{
        \begin{subarray}{l}
        t \sim \mathcal{U}[0, T] \\
        \mathbf{x}_t,y \sim p(\mathbf{x}_t,y) \\
        \end{subarray}
        }\left[\lambda(t) \|\nabla\log p(\mathbf{x}_t|y) - \mathbf{s}_\phi(\mathbf{x}_t,t,y)\|^2 \right]~.
$$
\begin{proof}
Because $\mathbf{x}_t$ is conditionally independent of $y$ given $\mathbf{x}_0$, we have
    \begin{equation}
    \label{eq:proof_1}
    \begin{aligned}
    ~&\mathbb{E}_{
        \begin{subarray}{l}
        t \sim \mathcal{U}[0, T] \\
        \mathbf{x}_0,y \sim p(\mathbf{x}_0,y) \\
        \mathbf{x}_t \sim p(\mathbf{x}_t|\mathbf{x}_0)
        \end{subarray}
        }\left[\lambda(t) \|\nabla\log p(\mathbf{x}_t|\mathbf{x}_0) - \mathbf{s}_\phi(\mathbf{x}_t,t,y)\|^2 \right] \\
    &= \mathbb{E}_{
        \begin{subarray}{l}
        t \sim \mathcal{U}[0, T] \\
        y \sim p(y) \\
        \mathbf{x}_0 \sim p(\mathbf{x}_0|y)\\
        \mathbf{x}_t \sim p(\mathbf{x}_t|\mathbf{x}_0)
        \end{subarray}
        }\left[\lambda(t) \|\nabla\log p(\mathbf{x}_t|\mathbf{x}_0) - \mathbf{s}_\phi(\mathbf{x}_t,t,y)\|^2 \right] \\
    &= \mathbb{E}_{
        \begin{subarray}{l}
        t \sim \mathcal{U}[0, T] \\
        y \sim p(y) \\
        \mathbf{x}_0 \sim p(\mathbf{x}_0|y)\\
        \mathbf{x}_t \sim p(\mathbf{x}_t|\mathbf{x}_0,y)
        \end{subarray}
        }\left[\lambda(t) \|\nabla\log p(\mathbf{x}_t|\mathbf{x}_0,y) - \mathbf{s}_\phi(\mathbf{x}_t,t,y)\|^2 \right] \\
    &= \mathbb{E}_{
        \begin{subarray}{l}
        t \sim \mathcal{U}[0, T] \\
        y \sim p(y) \\
        \end{subarray}
        }\left[\Phi(t,y)\right]~,
    \end{aligned} 
    \end{equation}
where 
$$
\begin{aligned}
    ~&\Phi(t,y)\\
    &=\mathbb{E}_{
        \begin{subarray}{l}
        \mathbf{x}_0 \sim p(\mathbf{x}_0|y)\\
        \mathbf{x}_t \sim p(\mathbf{x}_t|\mathbf{x}_0,y)
        \end{subarray}
        }
        \left[\lambda(t) \|\nabla\log p(\mathbf{x}_t|\mathbf{x}_0,y) - \mathbf{s}_\phi(\mathbf{x}_t,t,y)\|^2 \right]~.
\end{aligned}
$$
    
If $y$ and $t$ are fixed, we can define a transition probability not depend on these variables, $q(\mathbf{x}_0) = p(\mathbf{x}_0|y)$, $\kappa(\mathbf{x}_t)=\mathbf{s}_\phi(\mathbf{x}_t,t,y)$. According to~\cite{vincent2011connection}, we have:
\begin{equation}
\label{eq:proof_2}
\begin{aligned}
    \Phi(t,y) &= \mathbb{E}_{
        \begin{subarray}{l}
        \mathbf{x}_0 \sim q(\mathbf{x}_0)\\
        \mathbf{x}_t \sim q(\mathbf{x}_t|\mathbf{x}_0)
        \end{subarray}
        }
        \left[\lambda(t) \|\nabla\log q(\mathbf{x}_t|\mathbf{x}_0) - \kappa(\mathbf{x}_t)\|^2 \right] \\
    &= \mathbb{E}_{
        \begin{subarray}{l}
        (\mathbf{x}_0,\mathbf{x}_t) \sim q(\mathbf{x}_0,\mathbf{x}_t)\\
        \end{subarray}
        }
        \left[\lambda(t) \|\nabla\log q(\mathbf{x}_t|\mathbf{x}_0) - \kappa(\mathbf{x}_t)\|^2 \right] \\
    &= \mathbb{E}_{
        \begin{subarray}{l}
        \mathbf{x}_t \sim q(\mathbf{x}_t)\\
        \end{subarray}
        }
        \left[\lambda(t) \|\nabla\log q(\mathbf{x}_t) - \kappa(\mathbf{x}_t)\|^2 \right] \\
     &= \mathbb{E}_{
        \begin{subarray}{l}
        \mathbf{x}_t \sim p(\mathbf{x}_t|y)\\
        \end{subarray}
        }
        \left[\lambda(t) \|\nabla\log p(\mathbf{x}_t|y) - \mathbf{s}_\phi(\mathbf{x}_t,t,y)\|^2 \right]~.
    % &= \mathbb{E}_{
    %     \begin{subarray}{l}
    %     \mathbf{x}_t \sim q(\mathbf{x}_t)\\
    %     \end{subarray}
    %     }
    %     \left[\lambda(t) \|\nabla\log q(\mathbf{x}_t) - \kappa(\mathbf{x}_t)\|^2 \right] \\
\end{aligned}
\end{equation}

From Equation~\ref{eq:proof_1} and~\ref{eq:proof_2}, we can prove the equivalence of the two objective functions.

 \begin{equation}
    \begin{aligned}
    ~&\mathbb{E}_{
        \begin{subarray}{l}
        t \sim \mathcal{U}[0, T] \\
        \mathbf{x}_0,y \sim p(\mathbf{x}_0,y) \\
        \mathbf{x}_t \sim p(\mathbf{x}_t|\mathbf{x}_0)
        \end{subarray}
        }\left[\lambda(t) \|\nabla\log p(\mathbf{x}_t|\mathbf{x}_0) - \mathbf{s}_\phi(\mathbf{x}_t,t,y)\|^2 \right] \\
    &= \mathbb{E}_{
        \begin{subarray}{l}
        t \sim \mathcal{U}[0, T] \\
        y \sim p(y) \\
        \mathbf{x}_t \sim p(\mathbf{x}_t|y)
        \end{subarray}
        }\left[\lambda(t) \|\nabla\log p(\mathbf{x}_t|y) - \mathbf{s}_\phi(\mathbf{x}_t,t,y)\|^2 \right] \\
    &= \mathbb{E}_{
        \begin{subarray}{l}
        t \sim \mathcal{U}[0, T] \\
        (\mathbf{x}_t,y) \sim p(\mathbf{x}_t,y) \\
        \end{subarray}
        }\left[\lambda(t) \|\nabla\log p(\mathbf{x}_t|y) - \mathbf{s}_\phi(\mathbf{x}_t,t,y)\|^2 \right]~.
    \end{aligned} 
    \end{equation}
\end{proof}
% After training the score network model, we can replace the $\nabla\log p(\mathbf{x}_t|y)$ term in Equation~\ref{eq:PF_ODE} with a neural network:
% \begin{equation}
%      \frac{d\mathbf{x}_t}{dt} = -\frac{1}{2}\gamma_t\left[\mathbf{x}_t + \mathbf{s}_\phi(\mathbf{x}_t,t,y)\right]
% \end{equation}

\textbf{Discretization.} Consider discretizing the time horizon $[\epsilon,T]$ into $N-1$ with boundary $t_1=\epsilon<t_2<t_3<\ldots<t_{N}=T$. If $N$ is sufficiently large, we can use an ODE-solver~\cite{gupta1985review} to estimate the next discretization step:

\begin{equation}
    \label{eq:euler_solver}
    \begin{aligned}
        \hat{\mathbf{x}}_{t_i} &= \mathbf{x}_{t_{i+1}} + (t_i - t_{i+1}) \left. \frac{d\mathbf{x}}{dt} \right|_{t = t_{i+1}} \\
        &= \mathbf{x}_{t_{i+1}} - \frac{1}{2}\gamma_{i+1} (t_i - t_{i+1})\left[\mathbf{x}_{t_{i+1}} + \mathbf{s}_\phi(\mathbf{x}_t,t,y)\right]~.
    \end{aligned}
\end{equation}

\textbf{Conditional Consistency Model Loss.} To enable fast sampling, we expect that the predicted point $\hat{\mathbf{x}}_{t_i}$ and $\mathbf{x}_{t_{i+1}}$ to lie on the same probability flow ODE trajectory. We propose conditional consistency lost to enforce this constraint:
\begin{equation}
    \begin{aligned}
        \mathcal{L}_{\rm consistency} &= \mathbb{E}_{
        \begin{subarray}{l}
        i \sim \mathcal{U}[1, N - 1] \\
        \mathbf{x}_{t_{i+1}} \sim p(\mathbf{x}_{t_{i+1}}|\mathbf{x}_0)
        \end{subarray}
        } \\
        &\left[\lambda(t_i) \|\mathbf{f}_\theta(\mathbf{x}_{t_{i+1}},t_{i+1},y) - \mathbf{f}_{\theta^*}(\hat{\mathbf{x}}_{t_{i}},t_{i},y)\|^2 \right]~,
    \end{aligned}
\end{equation}
where $\hat{\mathbf{x}}_{t_i}$ is calculated in Equation~\ref{eq:euler_solver}, $\mathbf{x}_{t_{i+1}}$ is sampling from Gaussian distribution in Equation~\ref{eq:distribution_gaussian}, $\theta$ is parameters of neural network $\mathbf{f}$.

Additionally, we need to minimize the discrepancy between the predicted and ground truth grasp poses with the detection loss:
\begin{equation}
    \mathcal{L}_{\rm detection} = \mathbb{E}_{
        \begin{subarray}{l}
        i \sim \mathcal{U}[1, N] \\
        \mathbf{x}_{t_{i}} \sim \mathcal{N}(\mu_{t_{i}},\Sigma_{t_{i}}) \\
        \mathbf{x}_0,y \sim p(\mathbf{x}_0,y)
        \end{subarray} 
        }\left[\lambda(t_i)\|\mathbf{f}_\theta(\mathbf{x}_{t_i}, t_i, y) - \mathbf{x}_0\|^2\right]~.
\end{equation}

The overall training objective for our method is:

\begin{equation}
    \mathcal{L}_{\rm total} = \mathcal{L}_{\rm consistency} + \mathcal{L}_{\rm detection}~.
\end{equation}

\subsection{Network Details}
The input of our network is the image and a corresponding grasping text prompt represented as $e$ (for example, ``grasp the fork at its handle"). We first extract the image feature using a 12-layer vision transformer ViT~\cite{dosovitskiy2020image} image encoder. The input text prompt is encoded by a text encoder using BERT~\cite{devlin2018bert} or CLIP~\cite{radford2021learning}. We then combine and learn the features of the input text prompt and input image using the ALBEF fusion network~\cite{li2021align}. The output of the fusion features is fed into a score network and our conditional consistency model to learn the grasp pose. Figure~\ref{fig:lightgrasping} shows the detail of our network.

\textbf{Score Network.} 
In practice, we utilize a score network composed of several MLP layers to extract three components: the noisy grasp pose $\mathbf{x}_t$, the time index $t$, and the conditional vision-language embedding $y$. Subsequently, these features are concatenated and the score function is extracted through a 
final MLP layer. It is crucial to ensure that the output dimension of the score network is identical to the dimension of the input $\mathbf{x}_t$ because, fundamentally, the score function is the gradient of the grasp pose distribution given the condition $y$. Our conditional consistency model's network has an architecture similar to the score network; however, its output is the predicted grasp pose.

\vspace{2ex}
\begin{algorithm}[H]
\SetAlgoLined
 \KwIn{Image and text prompt, conditional consistency model $\mathbf{f}_\theta(\mathbf{x},t,y)$, number of inference step $P$, sequence of time points $t_1 = \epsilon < t_2 < t_3 < \dots < t_{P} = T$, noise scheduler $\alpha_t = e^{\rho_t}$.}
 $y \gets \text{ALBEF (image, prompt)}$\\Initial grasp noise $\mathbf{x}_T \sim \mathcal{N}(0,\mathbf{I})$\\
 $\mathbf{x}_0 \gets \mathbf{f}_\theta(\mathbf{x}_T,T,y)$\\
 \For{$i = P - 1$ \KwTo $2$}{
 Sample $\mathbf{z}~\sim \mathcal{N}(0,\mathbf{I})$\\
 $\mathbf{x}_{t_i} \gets \sqrt{\alpha_{t_i}}\mathbf{x}_0 + \sqrt{1 - \alpha_{t_i}}\mathbf{z}$\\
 $\mathbf{x}_0 \gets \mathbf{f}_\theta(\mathbf{x}_{t_i},t_i,y)$
 }
 \KwOut{Final grasp pose $\mathbf{x}_0$}
 
 \caption{Inference Process}
 \label{algo:multistep_inference}
\end{algorithm}

\subsection{Training and Inference}
During the training, we freeze the text encoder and image encoder, and then train the ALBEF fusion, the score network, and the consistency model end-to-end. We note that the score network and the conditional consistency model share the same architecture. We trained both models simultaneously for 1000 epochs with a batch size of 8 using Adam optimizer. The training time takes approximately 3 days on an NVIDIA A100 GPU. Regarding the parameters of the conditional consistency model, we empirically set $T = 1000$, $\epsilon = 1$, and $N = 2000$. %We follow Karras \textit{et al.}~\cite{karras2022elucidating} to determine the boundary time points with the formula $t_i = \left(\epsilon^{\frac{1}{\rho}} + \frac{i-1}{N-1}(T^{\frac{1}{\rho}} - \epsilon^{\frac{1}{\rho}})\right)^\rho$, with $\rho$ is set to 7. 
After training the score network and the conditional consistency model $\mathbf{f}_\theta(\mathbf{x}_t,t,y)$, we can sample the grasp pose given the input image and language instruction prompt in a few denoising steps using our  Algorithm~\ref{algo:multistep_inference}.

\section{Experiments} \label{Sec:exp}
% \lipsum[1]

\subsection{Experiment Setup}

\textbf{Dataset.} We use the Grasp-Anything dataset~\cite{vuong2023grasp} in our experiment. Grasp-Anything is a large-scale dataset for language-driven grasp detection with 1M samples. Each image in the dataset is accompanied by one or several prompts describing a general object grasping action or grasping an object at a specific location. %For example, \textit{``Grasp me the spoon"}, \textit{``Pick up the spoon at its handle"}. 
%As in~\cite{vuong2023grasp}, we divide the dataset into two sets, seen and unseen, to verify the robustness of our method. 

%Additionally, we are using images from other datasets~\cite{cao2023nbmod,xiang2017posecnn, fang1billion}.

\textbf{Evaluation Metrics.}
Our primary evaluation metric is the success rate, defined similarly to~\cite{kumra2020antipodal,vuong2023grasp}, necessitating an IoU score of the predicted grasp exceeding $25\%$ with the ground truth grasp and an offset angle less than $30^\circ$. We also use the harmonic mean (`H') to measure the overall success rates as in~\cite{zhou2022conditional}. The latency (inference time) in seconds of all methods is reported using the same NVIDIA A100 GPU.

\subsection{Comparison with Grasp Detection Methods}

\begin{table}[ht]
\caption{Comparision with Traditional Grasp Detection Methods}
\label{tab: zero-shot task result}
\vskip 0.15in
\setlength{\tabcolsep}{5.5pt}
\begin{center}
\begin{tabular}{llccc}
\toprule
Baseline & Seen  & Unseen & H & Latency\\
\midrule
GR-ConvNet~\cite{kumra2020antipodal} & 0.37 & 0.18 & 0.24 & \textbf{0.022}\\
Det-Seg-Refine~\cite{ainetter2021end} & 0.30 & 0.15 & 0.20 & 0.200 \\
GG-CNN~\cite{morrison2018closing} & 0.12 & 0.08 & 0.10 & 0.040\\
CLIPORT~\cite{shridhar2022cliport} & 0.36 & 0.26 & 0.29 & 0.131\\
CLIP-Fusion~\cite{xu2023joint} & 0.40 & 0.29 & 0.33 & 0.157\\
MaskGrasp~\cite{tuan2024seggrasp} & 0.50 & 0.46 & 0.45 & 0.116\\
%\midrule
%LGD + BERT & 0.44 & 0.38 & 0.41 & 25.36s \\
%LGD + CLIP & 0.48 & \bf{0.42} & 0.45 & 26.12s \\
\midrule

LLGD (ours) with 1 timestep & 0.47 & 0.34 & 0.40 & 0.035 \\
LLGD (ours) with 3 timesteps & 0.52 & 0.38 & 0.45 & 0.106 \\
LLGD (ours) with 10 timesteps & \textbf{0.53} & \textbf{0.39} & \textbf{0.46} & 0.264 \\
%LLGD (ours) with 30 timesteps & 0.54 & 0.40 & 0.47 & 0.763 \\
\bottomrule
\end{tabular}
\end{center}
\vskip -0.1in
\end{table}

We compare our LLGD with GR-CNN~\cite{kumra2020antipodal}, Det-Seg-Refine~\cite{ainetter2021end}, GG-CNN~\cite{morrison2018closing}, CLIPORT~\cite{shridhar2022cliport}, MaskGrasp~\cite{tuan2024seggrasp}, and CLIP-Fusion~\cite{xu2023joint}. %For a fair comparison, with GR-CNN, Det-Seg-Refine, and GG-CNN, we add the text encoder using CLIP~\cite{radford2021learning} and also utilize ALBEF~\cite{li2021align} to do the fusion between the text embedding and the visual features. For CLIPORT and CLIP-Fusion, we modify the last layer to output the rectangle grasp pose. The remaining training procedure is inherited from the original works. 
%In all cases, we employ a pretrained CLIP~\cite{radford2021learning} or BERT~\cite{devlin2018bert} as the text embedding. 
Table~\ref{tab: zero-shot task result} compares our method and other baselines on the GraspAnything dataset. This table shows that our proposed LLGD outperforms traditional grasp detection methods by a clear margin. Our inference time is also competitive with other methods. %Furthermore, Table~\ref{tab: zero-shot task result} shows that our LLGD archives reasonable results with just a few denoising steps, which significantly reduce the latency during inference.

%using CLIP~\cite{radford2021learning} as the text encoder brings slightly better results than BERT~\cite{devlin2018bert}.

%The findings indicate a notable performance advantage of our method over other baseline approaches. Our results outperform all other baseline models that use direct grasp detection from image and text prompts. Best overall success-rate in our method surpass CLIP-Fusion~\cite{} about 13\%, is a significantly improved result. When compared to LGD~\cite{}, a Diffusion-based model, the latency of our model utilizing the Conditional Consistency Model is approximately 100 times faster while still yielding equivalently good results. Our model leverages the accuracy of the Diffusion Model while achieving the fast inference time like direct real-time models from features. Figure~\ref{fig:visualize_grasping} visualizes and compares the predictive results of our and other baselines model.
% In particular, on the full-shape task, OpenAD significantly surpasses the runner-up model (ZSLPC) by 4.40\% on mIoU. OpenAD also outperforms others on Acc (0.84\% over 3DGenZ) and on mAcc (0.81\% over ZSLPC). Similarly, for the partial-view task, our method has the highest mIoU (3.98\% higher than the second-best ZSLPC), and also the highest Acc and mAcc.

% \begin{figure}[ht]
%     \centering
%     \includegraphics[width=\linewidth]{figures/exp/results_lightgrasp.pdf}
%     \caption{\textbf{Light-Grasping results visualization.  }}
%     \label{fig:results_lightgrasp}
% \end{figure}

\subsection{Comparison with Lightweight Diffusion Models}

\begin{table}[ht]
\caption{Comparison with Diffusion Models for Language-Driven Grasp Detection}
\label{tab_lightweight_diffusion_comparision}
\vskip 0.15in
\setlength{\tabcolsep}{5pt}
\begin{center}
\begin{tabular}{llccc}
\toprule
Method & Seen  & Unseen & H & Latency\\
\midrule
LGD\cite{vuong2024language} with 3 timesteps & 0.42 & 0.29 & 0.35 & 0.074 \\
LGD\cite{vuong2024language} with 30 timesteps & 0.49 & 0.41 & 0.45 & 0.741 \\
LGD\cite{vuong2024language} with 1000 timesteps & 0.52 & 0.42 & 0.47 & 26.12 \\
SnapFusion~\cite{li2023snapfusion} with 500 timesteps& 0.49 & 0.37 & 0.43 & 12.95\\
LightGrad~\cite{chen2023lightgrad} with 250 timesteps & 0.51 & 0.34 & 0.43 & 6.420 \\ \midrule
LLGD (ours) with 1 timestep & 0.47 & 0.34 & 0.40 & \textbf{0.035} \\
LLGD (ours) with 3 timesteps & 0.52 & 0.38 & 0.45 & 0.106 \\
LLGD (ours) with 10 timesteps & \textbf{0.53} & \textbf{0.39} & \textbf{0.46} & 0.264 \\
\bottomrule
\end{tabular}
\end{center}
\vskip -0.1in
\end{table}
In this experiment, we compare our LLGD with other diffusion models for language-driven grasp detection. In particular, we compare with LGD\cite{vuong2024language} using DDPM~\cite{ho2020denoising}, and recent lightweight diffusion works: SnapFusion~\cite{li2023snapfusion} with 500 timesteps and LightGrad~\cite{chen2023lightgrad} with 250 timesteps. %. With SnapFusion, we use step distillation technique to reduce the total number of steps from 1000 to 500. We also use DPM-Solver-$k$~\cite{chen2023lightgrad} for LightGrad with 250 time points uniformly distributed in the range from 0 to 1000 without using knowledge distillation.

Table~\ref{tab_lightweight_diffusion_comparision} shows the result diffusion models for language-driven grasp detection. We can see that the accuracy and inference time of the classical diffusion model LGD strongly depend on the number of denoising timesteps. LGD with 1000 timesteps achieves reasonable accuracy but has a significant long latency. Lightweight diffusion models such as SnapFusion~\cite{li2023snapfusion} and LightGrad~\cite{chen2023lightgrad} show reasonable results and inference speed. However, our method achieves the highest accuracy with the fastest inference speed.

%method. In this experiment, we use DDPM~\cite{ho2020denoising} and two other fast sampling techniques in SnapFusion~\cite{li2023snapfusion} and LightGrad~\cite{chen2023lightgrad} to compare their effectiveness in terms of accuracy and latency against our proposed method. Li \textit{et al.}~\cite{li2023snapfusion} utilize step distillation~\cite{salimans2022progressive} to learn a student model with a half step compare with original model. This method can be applied to discrete diffusion models, but it requires multiple iterations to reduce the number of inference steps by factors of 2, 4, 8, \dots~ Chen \textit{et al.}~\cite{chen2023lightgrad} used the DPM-Solver~\cite{lu2022dpm} as an effective method for fast sampling in the continuous diffusion model. However, due to the absence of knowledge distillation, the quality of the model significantly decreased despite the fast inference time.

\subsection{Conditional Consistency Model Demonstration}
\begin{figure}[ht]
    \centering
    \includegraphics[width=\linewidth, height=0.5\linewidth]{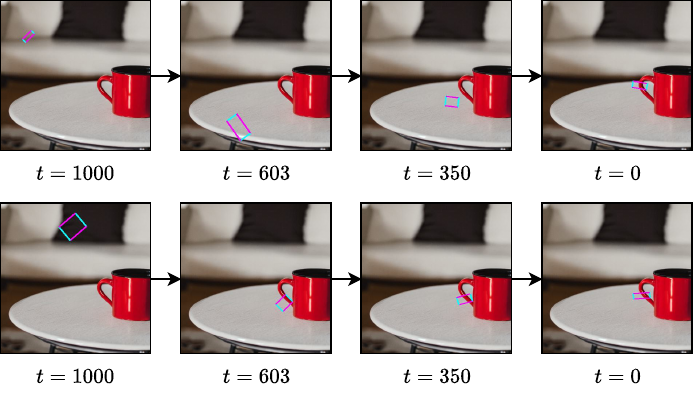}
    %\vspace{0.5mm}
    \caption{Consistency model analysis. With text prompt input \textit{``Grasp the cup at its handle"}, we compare the trajectory grasp pose of our method and LGD\cite{vuong2024language}. In the figure, the top row illustrates the trajectory of LGD, while the bottom row corresponds to the trajectory of our LLGD.}
    \label{fig:method_analysis}
\end{figure}

\begin{figure*}[ht]
\centering
\includegraphics[width=\linewidth]{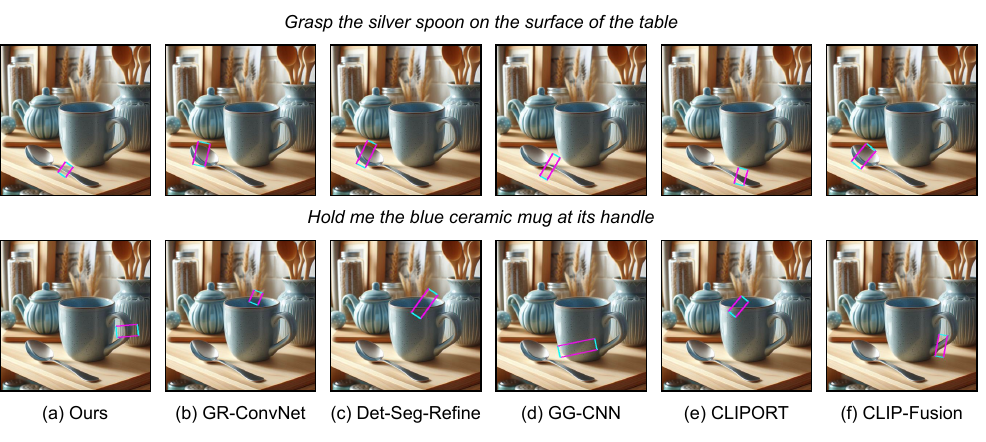} 
   %\vspace{-2ex}
   \caption{Visualization of detection results of different language-driven grasp detection methods.}
   \label{fig:visualize_grasping}
\end{figure*}

In this analysis, we will verify the effectiveness of our conditional consistency model. %Additionally, we compare trajectory grasp pose between our method and traditional diffusion using DDPM~\cite{ho2020denoising} with 1000 steps. 
In Figure~\ref{fig:method_analysis}, we visualize grasp pose aspect to time index $t$. In the LGD\cite{vuong2024language} model, as the discrete diffusion model is employed with
$T=1000$, we have to perform the diffusion steps with a step size of 1, which results in very slow inference speed. Moreover, the grasp pose trajectory still exhibits significant fluctuations. Our method can arbitrarily select boundary time points for the continuous consistency model. It is evident that the number of iterations required by our method is significantly less than that of LGD\cite{vuong2024language} for the same value of $T$, which contributes to the ``lightweight" factor. Furthermore, the grasp pose at $t=603$ has almost converged to the ground truth, while LGD\cite{vuong2024language} using DDPM at $t=350$ has not yet achieved a successful grasp. %This demonstrates our method's consistency and more stable compared to the traditional diffusion model. 

% \begin{table}[ht]
% \caption{Text Encoder Analysis}
% \label{tab_dependency_step}
% \vskip 0.15in
% \begin{center}
% \begin{tabular}{llccc}
% \toprule
% Method & Seen  & Unseen & H & Latency\\
% \midrule
% % LLGD with 1 steps & 0.47 & 0.34 & 0.40 & 0.035s \\
% % LLGD with 4 steps & 0.52 & 0.38 & 0.45 & 0.106s \\
% % LLGD with 10 steps & 0.53 & 0.39 & 0.46 & 0.264s \\
% % LLGD with 30 steps & 0.54 & 0.40 & 0.47 & 0.763s \\

% LLGD + BERT~\cite{devlin2018bert} & 0.48 & 0.35 & 0.41 & \textbf{0.099 }\\
% LLGD + CLIP~\cite{radford2021learning} & \bf{0.52} & \textbf{0.38} & \bf{0.45} & 0.106\\
% \bottomrule
% \end{tabular}
% \end{center}
% \vskip -0.1in
% \end{table}

\subsection{Ablation Study}

% \begin{figure}[ht]
%   \centering
%   % Scale hình vẽ nếu cần - sửa số 1.0 thành tỉ lệ thích hợp
%   \def\svgwidth{1.0\columnwidth}
%   \input{figures/robot_demonstration/robotic-experiment.pdf_tex}
%   \caption{Robotic systems for Light-Grasping}
%   \label{fig:robot_system}
% \end{figure}

%\textbf{Text Encoder Analysis.} Table~\ref{tab_dependency_step} highlights the efficacy of using different text encoders in our LLGD with 3 timesteps. We can see that CLIP outperforms BERT by a clear margin as CLIP learns visual concepts more effectively through text prompts. This enhanced capability comes at the cost of a slightly increased inference time (latency) in our method.

\textbf{Visualization.} Figure~\ref{fig:visualize_grasping} shows qualitative results of our method and other baselines. The outcomes suggest that our method LLGD generates more semantically plausible grasp poses given the same text query than other baselines. In particular, other methods usually show grasp poses at the location that is not well-aligned with the text query, while our method shows more suitable detection results.

%Our method has effectively leveraged the language understanding capabilities of the conditional consistency model, particularly for the task of grasping objects at specific position.

\begin{figure}[h]
    \centering
    \includegraphics[width=\linewidth]{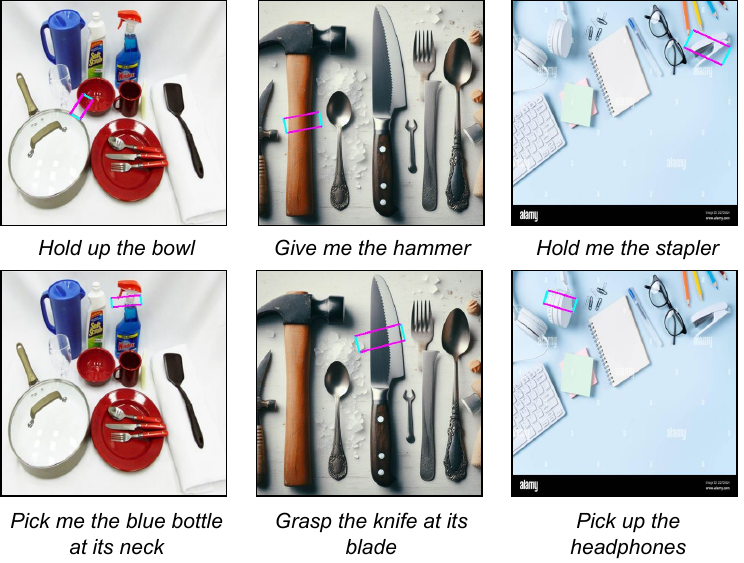}
    \vspace{-1ex}
    \caption{In the wild detection results. Images are from the internet. % The grasp poses identified are of satisfactory quality. This suggests a robust performance of LLGD on unseen data sources. 
    }
    \label{fig:in_the_wild_detection}
\end{figure}

\textbf{In the Wild Detection.} Figure~\ref{fig:in_the_wild_detection} illustrates the outcomes of applying our method to random images from the internet. The results demonstrate that our LLGD can effectively detect the grasp pose given the language instructions on real-world images. Our method showcases a promising zero-shot learning ability, as it successfully interprets grasp actions on images it has never encountered during training.

\begin{figure}[t]
    \centering
    \includegraphics[width=\linewidth]{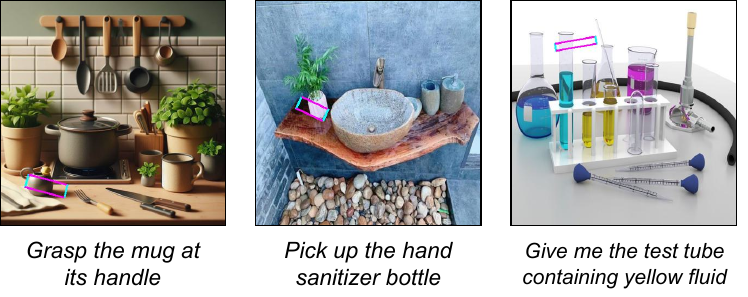}
    \vspace{-2ex}
    \caption{Prediction failure cases. % When objects have similar shapes, are occluded, are small in size, or are uncommon items such as \textit{``test tube"}, our method still yields incorrect predictions.
    }
    \label{fig:failure_sample}
\end{figure}

\textbf{Failure Cases.} Although good results have been achieved, our method still predicts incorrect grasp poses. A large number of objects and grasping prompts pose a challenging problem as the network cannot capture all the diverse circumstances that arise in real life. Figure~\ref{fig:failure_sample} depicts some failure cases where LLGD incorrectly predicts the results, which can be attributed to the presence of multiple similar objects that are difficult to distinguish and text prompts that lack detailed descriptions for accurate result determination.

\subsection{Robotic Experiments}

\textbf{Robotic Setup.} Our lightweight language-driven grasp detection pipeline is incorporated within a robotic grasping framework that employs a KUKA LBR iiwa R820 robot to deliver quantifiable outcomes. Utilization of the RealSense D435i camera enables the translation of grasping information from LLGD into a 6DoF grasp posture, bearing resemblance to~\cite{kumra2020antipodal}. Subsequently, a trajectory optimization planner~\cite{vu2023machine} is used to execute the grasping action. Experiments were conducted on a table surface for two scenarios: the single object scenario and the cluttered scene scenario, wherein various objects were placed to test each setup. Table~\ref{tab_robotic_results} shows the success rate of our method and other baseline models. We can see that our method outperforms other baselines in both single object and cluttered scenarios. Furthermore, our lightweight model allows rapid execution speed without sacrificing the visual grasp detection accuracy.

\begin{table}[t]
    \centering
    \caption{\label{table: real-robot-exp} Robotic language-driven grasp detection results}
    \vspace{2ex}
    \renewcommand
\tabcolsep{4pt}
\hspace{1ex}
    \begin{tabular}{@{}rcc@{}}
\toprule
Baseline & Single &  Cluttered\cr 
\midrule
GR-ConvNet~\cite{kumra2020antipodal} + CLIP~\cite{radford2021learning}  &0.33  & 0.30\\ 
Det-Seg-Refine~\cite{ainetter2021end} + CLIP~\cite{radford2021learning} &0.30  & 0.23\\ 
GG-CNN~\cite{morrison2018closing} + CLIP~\cite{radford2021learning} &0.10  & 0.07 \\
CLIPORT~\cite{shridhar2022cliport} &0.27 & 0.30 \\
CLIP-Fusion~\cite{xu2023joint} & 0.40 & 0.40 \\
SnapFusion~\cite{li2023snapfusion}& 0.40 & 0.39 \\
LightGrad~\cite{chen2023lightgrad}  & 0.41 & 0.40 \\
\midrule
LLGD (ours) &  \textbf{0.43} & \textbf{0.42} \\
\bottomrule
\end{tabular}
\label{tab_robotic_results}
\end{table}

%\begin{figure}[ht]
%    \centering
%    \includegraphics[width=\linewidth]{figures/exp/robotic_action.pdf}
%    \caption{\textbf{Illustrative snapshots from a robotic experiment.}}
%    \label{fig:robotic_action}
%\end{figure}

 %Figure~\ref{fig:robotic_action} illustrates a sequence of robot actions when executing a grasping action given a text query. 
%We note that our method successfully differentiates between objects of similar structure but varying properties, for example ``\textit{green pen}" and ``\textit{red pen}". In such cases, the robot needs to identify which are the objects to be grasped based on adjectives describing them if there are multiple similar objects in the image. 

% \begin{figure}[h]
% \centering
% \def\svgwidth{1\columnwidth}
% \input{robot-experiment-nghia.pdf_tex}
% %\vspace{1ex}
% \caption{Illustrative snapshots from a robotic experiment. The text input query is ``grasp the green pen".}
% \label{fig:robotic_action}
% \end{figure}

\section{Discussion}\label{Sec:con}

\textbf{Limitation. } Despite achieving notable results in real-time applications, our method still has limitations and predicts incorrect grasp poses in challenging real-world images. Faulty grasp poses are often due to the correlation between the text and the attention map of the visual features not being well-aligned as shown in Fig.~\ref{fig:failure_sample}. From our experiment, we see that when grasp instruction sentences contain rare and challenging nouns that are popular in the dataset, ambiguity in parsing or text prompts would happen and is usually the main cause that leads to the incorrect prediction of grasp poses. Therefore, providing the instruction prompts with clear meanings is essential for the robot to understand and execute the correct grasping action.
% Our approach is quite effective and fast when it comes to handling tasks in the real world, showing that it's pretty useful. However, there are certain limitations. 
% \begin{figure}
% \centering
% \def\svgwidth{1\columnwidth}
% \input{robotic-experiment.pdf_tex}
% \vspace{-2mm}
% \caption{Overview of the robotic experiment setup}
% \label{fig: robot demonstration}
% \end{figure}

\textbf{Future work.} 
%\section*{Acknowledgment}
%\addcontentsline{toc}{section}{Acknowledgment}
%\lipsum[1]
% From our intensive experiments, we
% see several improvement points for future work. First, future work could explore the use of lightweight models for processing 3D point clouds and depth images, addressing the information gap in scene understanding that is often present with 2D images. Additionally, we propose leveraging Graph Convolutional Networks (GCNs)~\cite{kipf2016semi} to recognize local geometric structures of similar objects and differentiate between them effectively. 
We see several prospects for improvement in future work: \textit{i)} expanding our method to handle 3D space is essential, implementing it for 3D point clouds and RGB-D images to avoid the lack of depth information in robotic applications, \textit{ii)} addressing the gap between the semantic
concept of text prompts and input image, analyzing the detailed geometry of objects for distinguishing between items with similar structure, and \textit{iii)} expanding the problem to more complex language-driven manipulation applications, for instance, in case the robots want to grasp a plate containing apples, the robot would need to manipulate the objects in such a manner that prevents the apples from falling.

%\section{Conclusion}
%We propose a new method for real-time language-driven grasp detection problems using a conditional consistency model. Unlike traditional approaches, our LLGD leverages the advantages of diffusion-based models in language-visual understanding. We tackle the slow inference speed of the diffusion models with a new conditional consistency model. The intensive experimental results show that our LLGD outperforms other methods by a clear margin while offering competitive execution times. We further verify the capability of our LLGD to detect grasp poses on both single and cluttered scenes in real robot experiments. Our source code and trained model will be released for further study. 

%. We additionally demonstrated the usability of LLGD in real-world robotic applications.

\bibliographystyle{class/IEEEtran}
\bibliography{class/IEEEabrv,class/reference}
   
\end{document}